\documentclass[runningheads]{llncs}
\usepackage[T1]{fontenc}
\usepackage{comment}
\usepackage{graphicx}
\usepackage{amsmath}
\usepackage{amssymb}
\usepackage{booktabs}   
\usepackage{graphicx}   
\usepackage{multirow}
\usepackage{wrapfig}
\usepackage{pifont}
\newcommand{\cmark}{\ding{51}}
\newcommand{\xmark}{\ding{55}}
\newcommand{\murim}{\texttt{MURIM}}

\usepackage{xcolor}
\usepackage{tikz}
\usetikzlibrary{arrows.meta,positioning,calc}
\usepackage{fontawesome5} 

\definecolor{clientblue}{RGB}{227,242,253}   
\definecolor{servergreen}{RGB}{232,245,233}  
\definecolor{accentblue}{RGB}{25,118,210}    
\definecolor{accentgreen}{RGB}{46,125,50}    

\usepackage{array}
\newcolumntype{P}[1]{>{\centering\arraybackslash}p{#1}}
\usepackage{makecell}
\usepackage{xcolor}

\begin{document}
\title{MURIM: Multidimensional Reputation-based Incentive Mechanism for Federated Learning}
%
%
\author{Sindhuja Madabushi  \and
Dawood Wasif \and
Jin-Hee Cho }

%
%
\institute{
Virginia Tech\\
\email{\{msindhuja,dawoodwasif,jicho\}@vt.edu}}

\maketitle              
\begin{abstract}
Federated Learning (FL) has emerged as a leading privacy-preserving machine learning paradigm, enabling participants to share model updates instead of raw data. However, FL continues to face key challenges, including weak client incentives, privacy risks, and resource constraints. Assessing client reliability is essential for fair incentive allocation and ensuring that each client’s data contributes meaningfully to the global model. To this end, we propose {\murim}, a \underline{MU}lti-dimensional \underline{R}eputation-based \underline{I}ncentive \underline{M}echanism that jointly considers client reliability, privacy, resource capacity, and fairness while preventing malicious or unreliable clients from earning undeserved rewards. {\murim} allocates incentives based on client contribution, latency, and reputation, supported by a reliability verification module. Extensive experiments on MNIST, FMNIST, and ADULT Income datasets demonstrate that {\murim} achieves up to 18\% improvement in fairness metrics, reduces privacy attack success rates by 5–9\%, and improves robustness against poisoning and noisy-gradient attacks by up to 85\% compared to state-of-the-art baselines. 
Overall, {\murim} effectively mitigates adversarial threats, promotes fair and truthful participation, and preserves stable model convergence across heterogeneous and dynamic federated settings.
\end{abstract}
\section{Introduction}
Federated Learning (FL) is a distributed machine learning (ML) paradigm in which multiple clients, such as mobile devices, edge servers, or organizations, collaboratively train a shared ML or Deep Learning (DL) model without exchanging raw data. In each round, a central server coordinates training by selecting a subset of clients, distributing the current global model, and aggregating locally trained updates, typically via weighted averaging, until convergence. As such, FL has become a leading privacy-preserving ML framework, widely adopted by both academia and industry. Despite its promise, FL continues to face major challenges related to client incentives, privacy preservation, resource constraints, and fairness. Sustaining client participation requires appropriate incentives, yet fairness issues from data heterogeneity are often addressed only at the outcome level, overlooking geometric imbalances in how diverse updates shape the global model. Although gradients, not raw data, are shared, they remain vulnerable to inference and poisoning attacks. Many clients also lack sufficient computational capacity for intensive local training, making truthful resource reporting essential. Prior work has explored reliability-, privacy-, or fairness-aware incentive mechanisms in isolation, but rewarding any single factor can induce harmful behaviors such as oversharing or misreporting, motivating a multidimensional mechanism that jointly integrates privacy, fairness, reliability, and resource constraints.

To address these challenges, we propose {\murim}, a Multidimensional Reputation-based Incentive Mechanism for FL. {\murim} jointly integrates four dimensions, including privacy, resource constraints, fairness, and reliability, to incentivize clients in a manner that is trustworthy, equitable, and sustainable. Our goal is to design a robust, lightweight mechanism that avoids costly hardware solutions such as Trusted Execution Environments (TEE) or blockchain, enhances representation of underrepresented client updates, and rewards truthful participants proportionally to their contributions. Our key contributions are as follows:
\begin{enumerate}
\item We design a reliability mechanism that detects truthful clients based on consistency in privacy budgets, resource reports, and training behavior, enabling fairness-aware and privacy-preserving participation.
\item We propose the \emph{Subspace Leverage Equalizer (SLE)}, a fairness metric enhancing geometric representation of underrepresented clients during aggregation.

\item We design a unified incentive framework ensuring individual rationality, budget balance, and truthfulness under heterogeneous client behaviors.

\item We conduct extensive evaluations across multiple datasets (MNIST, FMNIST, ADULT)~\cite{kohavi1996adult,lecun1998mnist,xiao2017fmnist} and both IID and non-IID settings, demonstrating robust liar detection, stable convergence, and fairness improvements.
\end{enumerate}
Collectively, these contributions fill key gaps in existing incentive mechanisms (Section~\ref{sec:related-work}), which lack unified support for reliability checking, geometric fairness, truthful reward allocation, and rigorous evaluation under heterogeneous and adversarial FL settings. Details of prior IMs are discussed below.

\section{Related Work: Incentive Mechanisms (IMs) for FL}
\label{sec:related-work}

\begin{table}[t!]
\caption{\textbf{FL Algorithms on Privacy, Fairness, Truthfulness, and Reputation}}
\label{tab:algo_comparison}
\centering
\scriptsize
\begin{tabular}{|l|c|c|c|l|}
\hline
\textbf{HFL Scheme} & \textbf{Privacy} & \textbf{Fairness} & \textbf{Truthfulness} & \textbf{Reputation check method} \\ \hline
RRAFL\cite{zhang2021rrafl}        & \xmark & \xmark & \cmark & Indirect + moving average \\ \hline
Herath et al.\cite{herath2025lightweight} & \xmark & \cmark & \xmark & Peer-evaluated tables \\ \hline
RepAvg \cite{barkatsa2025repavg}       & \xmark & \cmark & \xmark & Long-term Shapley value \\ \hline
IMFGR \cite{ding2024imfgr}        & \xmark & \cmark & \cmark & Contribution + auction grade \\ \hline
TFFL \cite{rashid2025tffl}          & \xmark & \cmark & \xmark & Subjective logic consensus \\ \hline
RAIM \cite{zuo2023raim}         & \xmark & \xmark & \xmark & Blockchain history \\ \hline
RBFF \cite{song2022rbff}         & \xmark & \cmark & \xmark & Beta-distribution update \\ \hline
RIFL \cite{tang2024rifl}        & \xmark & \cmark & \xmark & Non-iid robust aggregation \\ \hline
Yang et al.\cite{yang2025fairnessaware} & \cmark & \cmark & \xmark & Stackelberg Game Rewards \\ \hline
FIM-FAN \cite{zhao2025fimfan}    & \xmark & \cmark & \cmark & Referral reverse auction \\ \hline
FIFL \cite{gao2025fifl}        & \cmark & \xmark & \cmark & Blockchain \& Rewards \\ \hline
FedFAIM \cite{shi2025fedfaim}     & \xmark & \cmark & \xmark & Shapley + Reputation \\ \hline
FDFL \cite{chen2024fdfi}      & \xmark & \cmark & \cmark & Provable Reward Allocation \\ \hline
FAR-AFL \cite{tang2025farafl}    & \xmark & \cmark & \cmark & Reverse Auction \\ \hline
IAFL \cite{wu2024incentive}      & \xmark & \cmark & \xmark & Contribution function agnostic rewards \\ \hline

Liu et al.\cite{liu2024eeppim}       & \cmark & \xmark & \xmark & Differential Privacy (DP) \\ \hline
Liu et al.\cite{liu2021picel}     & \cmark & \xmark & \xmark & Differential Privacy (DP) \\ \hline
PrivAim \cite{wang2023privaim}    & \cmark & \xmark & \xmark & Dual-Differential Privacy \\ \hline
Pain-FL \cite{sun2021painfl}      & \cmark & \xmark & \cmark & Personalized Differential Privacy \\ \hline
NICE \cite{xu2022nice}            & \cmark & \xmark & \cmark & Differential Privacy (DP) \\ \hline
CSRA \cite{li2024csra}             & \cmark & \xmark & \xmark & Differential Privacy (DP) \\ \hline
Aerial FL \cite{wang2024aerialfl}   & \cmark & \xmark & \cmark & DP, Encryption, Secure aggregation \\ \hline
IMPP-FL \cite{siqin2025auction}     & \cmark & \xmark & \cmark & DP, Encryption, Secure aggregation \\ \hline
SPNE \cite{mao2024game}             & \cmark & \xmark & \cmark & DP, Encryption, Secure aggregation \\ \hline

LCEME \cite{zhao2022lceml}              & \xmark & \xmark & \cmark & Reward based on effort/loss \\ \hline
FACT \cite{bornstein2024fact}                             & \xmark & \cmark & \cmark & Sandwich penalization \\ \hline
MODA \cite{jiang2024moda}              & \xmark & \cmark & \cmark & Double auction \\ \hline
FedAB \cite{wu2023fedab}                          & \xmark & \cmark & \cmark & Reverse auction + critical bid \\ \hline
Federated Unlearning \cite{xie2024forget} & \xmark & \xmark & \cmark & Data forgetting/zero-payment \\ \hline
FGCML\cite{xu2021gradient} & \xmark & \cmark & \xmark & Training-time incentives \\ \hline
\textbf{MUIRM (ours)} & \cmark & \cmark & \cmark & Data forgetting/zero-payment \\ \hline
\end{tabular}
\end{table}

\textbf{Fairness-aware IMs} uplift disadvantaged clients via contribution-based rewards~\cite{shi2025fedfaim,chen2024fdfl}, use privacy-preserving incentives such as differential privacy (DP), homomorphic encryption (HE), or secret sharing~\cite{yang2025fairnessaware}, and encourage truthful reporting through incentive-compatible designs~\cite{chen2024fdfl,zhao2025fimfan}. Yet existing schemes rarely unify fairness, privacy, and truthfulness; our approach integrates all three.

\textbf{Privacy-preserving IMs} include game-theoretic and auction-based designs~\cite{li2024csra,mao2024game,siqin2025auction,xu2022nice}, DP or HE–based schemes~\cite{liu2021picel,sun2021painfl,wang2023privaim,wang2024aerialfl,xu2022nice}, and contribution- or quality-aware fairness methods~\cite{chen2024fdfl,liu2021picel,liu2024eeppim,shi2025fedfaim,wang2023privaim}. Training-time and resource-aware incentives~\cite{liu2024eeppim,wang2024aerialfl,wu2024incentive} and TEE- or proof-based mechanisms~\cite{chen2024fdfl,zhao2025fimfan} further enhance overall reliability. Our work provides privacy-preserving incentives while jointly ensuring fairness and truthfulness.

\textbf{Truthful IMs} ensure honest cost or effort reporting via reward functions~\cite{zhao2022lceml}, penalization mechanisms~\cite{bornstein2024fact}, and auction designs such as double or reverse auctions~\cite{jiang2024moda,wu2023fedab}. Some works extend truthfulness to federated unlearning through zero-payment schemes~\cite{xie2024forget}. However, these methods typically secure incentive compatibility without providing privacy or fairness guarantees.

Table~\ref{tab:algo_comparison} summarizes existing IMs in horizontal FL (HFL) and compares them with our proposed {\murim}.  We adopt the following operational definitions: fairness is the measurable uplift for clients with low-quality or underrepresented data; truthfulness is the ability to verify client self-reports or guard against strategic misreporting; and privacy refers to protections using DP, HE, or secure multi-party computation (SMPC).

\section{Network and Threat Models}
\label{sec:network-threat-model}

\noindent \textbf{Network Model.}
We consider a synchronous horizontal FL system with a central server and $N \geq 2$ clients. Each round, the server broadcasts the global model, clients train locally for fixed epochs, add Gaussian noise for DP, and return noisy updates. The server aggregates these updates (Section~\ref{sec:proposed-approach}) to form the next global model. Beyond aggregation, the server computes incentives using a Subjective Logic (SL)~\cite{josang2016subjective}-based reliability framework. SL models uncertainty through \emph{belief}, \emph{disbelief}, and \emph{uncertainty} masses, enabling principled fusion of evidence across rounds. In our design, these opinions capture consistency in each client's privacy behavior and latency, allowing the server to derive a reliability score (i.e., reputation) that governs proportional reward allocation.

\noindent \textbf{Threat Model.}
We consider four representative attacks. \textbf{(a) Membership Inference Attack (MIA)} determines whether a sample belongs to a client’s training set by exploiting disparities in losses or confidence scores~\cite{hu2022membership}. \textbf{(b) Property Inference Attack (PIA)} detects whether a group-level attribute exists in a client’s data using batch-level logits or gradient patterns~\cite{melis2019exploiting}. \textbf{(c) Model Poisoning Attack (MPA)} inverts and scales a client’s update (e.g., $g_i^{\text{adv}}=-\gamma g_i$) to divert the global model from the optimum~\cite{fang2020local}. \textbf{(d) Noisy Gradient Attack (NGA)} injects high-variance noise (e.g., $g_i^{\text{adv}}=g_i+\sigma\mathcal{N}(0,I)$) to reduce the signal-to-noise ratio and destabilize convergence~\cite{liu2022friendly}.

\section{Proposed Approach: {\murim}}
\label{sec:proposed-approach}
\textbf{Contribution Evaluation.}
We compute the contribution of client $i$ in round $t$ by projecting its update $\boldsymbol{\Delta}_{i,t}$ onto the global update direction $\mathbf{v}$ and weighting it by the cosine similarity:
\begin{equation}
C_{i,t} = \boldsymbol{\Delta}_{i,t}\, \cos\theta_{i,t}\, |\cos\theta_{i,t}|,
\qquad
\cos\theta_{i,t} = 
\frac{\boldsymbol{\Delta}_{i,t} \cdot \mathbf{v}}
{\|\boldsymbol{\Delta}_{i,t}\|_2 \, \|\mathbf{v}\|_2}.
\end{equation}
Here, $\cos\theta_{i,t}$ is the cosine similarity between the local and global updates, and $C_{i,t}$ denotes the resulting contribution score.

\noindent \textbf{Resource--Latency Relationship.}
Following~\cite{schlegel2023codedpaddedfl}, the expected latency of client $i$ in round $t$ is modeled as
\begin{equation}
L_{i,t}^{\text{exp}} = \frac{\rho_i}{R_{i,t}^{\text{rep}}} + \Lambda_i,
\end{equation}
where $\rho_i$ is the MAC cost of one local update, $R_{i,t}^{\text{rep}}$ denotes the client’s reported computational resources, and $\Lambda_i$ is a fixed per-round overhead.


\noindent \textbf{Subspace Leverage Equalizer (SLE).}
SLE amplifies underrepresented update directions to prevent domination by a few client clusters. Let $\mathbf{g}_i$ be client $i$’s update and $\mathbf{u}_i=\mathbf{g}_i/\|\mathbf{g}_i\|_2$ its normalized direction. Stacking these into $U$, the ridge leverage score, weight, and aggregated update are jointly given by
\begin{equation}
\label{eq:sle}
\ell_i(\lambda)=\mathbf{u}_i^\top (U^\top U + \lambda I)^{-1}\mathbf{u}_i,\quad
\omega_i=\frac{\ell_i}{\sum_j \ell_j},\quad
\Delta\mathbf{w}_{\mathrm{SLE}}=\sum_{i=1}^n \omega_i\,\mathbf{g}_i.
\end{equation}
SLE boosts rare update directions to improve representational fairness.

\noindent \textbf{Reputation-based Compliance Evaluation.}
During \murim{} training, each client is assigned a \emph{privacy-budget interval} $[\varepsilon_{\min},\varepsilon_{\max}]$ specifying the allowable privacy levels the server expects when updates are perturbed.

\begin{definition}[Reliable Client]
\label{def:reliable_client}
A client $i$ is reliable in round $t$ if its privacy budget satisfies $\varepsilon_{\min} \le \varepsilon_{i,t} \le \varepsilon_{\max}$ and its observed latency $L_{i,t}$ is consistent with the expected latency $L_{i,t}^{\text{exp}}$ implied by its reported resources $R_{i,t}^{\text{rep}}$. Reliability therefore requires valid privacy behavior and latency aligned with claimed resources.
\end{definition}

\begin{definition}[Unreliable Client]
\label{def:unreliable_client}
A client $i$ is unreliable in round $t$ if it violates the privacy-budget interval ($\varepsilon_{i,t} < \varepsilon_{\min}$) or if its latency greatly exceeds the expected value ($L_{i,t} \gg L_{i,t}^{\text{exp}}$), indicating misreported resources.
\end{definition}


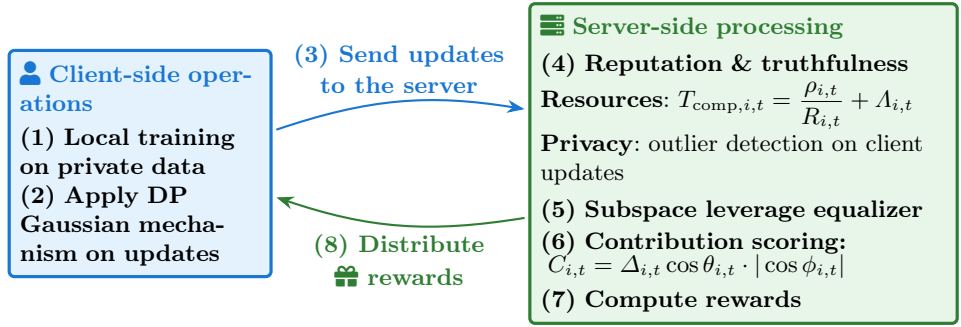
\begin{figure}[t]
  \centering
  \begin{tikzpicture}[
      font=\small,
      block/.style={
        draw,
        rounded corners=2pt,
        align=left,
        inner sep=4pt,
        outer sep=2pt,
      },
      clientblock/.style={
        block,
        fill=clientblue,
        draw=accentblue,
        very thick,
      },
      serverblock/.style={
        block,
        fill=servergreen,
        draw=accentgreen,
        very thick,
      },
      arrowCS/.style={-{Stealth}, thick, draw=accentblue},
      arrowSC/.style={-{Stealth}, thick, draw=accentgreen},
    ]

    \node[clientblock, text width=3.2cm] (client) {
      {\bfseries \textcolor{accentblue}{\faUser\ Client-side operations}}\\[2pt]
      \textbf{(1) Local training on private data}\\
      \textbf{(2) Apply DP Gaussian mechanism on updates}
    };

    \node[serverblock, right=3.3cm of client, align=left, text width=5.4cm] (server) {
      {\bfseries \textcolor{accentgreen}{\faServer\ Server-side processing}}\\[3pt]
      \shortstack{\textbf{(4) Reputation \& truthfulness}} \\[1pt]
      \textit{\bf Resources}: 
      $T_{\mathrm{comp},i,t}
        = \dfrac{\rho_{i,t}}{R_{i,t}} + \Lambda_{i,t}$\\[1pt]
      \textit{\bf Privacy}: outlier detection on client updates \\[3pt]
      \textbf{(5) Subspace leverage equalizer} \\[1pt]
      \textbf{(6) Contribution scoring:}\\[-2pt]
      \hspace*{0.3em}$C_{i,t} = \Delta_{i,t}\cos\theta_{i,t}\cdot|\cos\phi_{i,t}|$\\[2pt]
      \textbf{(7) Compute rewards}
    };

    \draw[arrowCS]
      (client)
        to[bend left=14]
      node[above]{\small \textbf{\textcolor{accentblue}{\shortstack{(3) Send updates \\ to the server}}}}
      (server);

    \draw[arrowSC]
      (server)
        to[bend left=14]
      node[below]{\small \textbf{\textcolor{accentgreen}{\shortstack{(8) Distribute \\ \faGift\ rewards}}}}
      (client);

  \end{tikzpicture}
  \caption{Overview of the proposed MURIM framework illustrating the end-to-end federated learning workflow.}
  \label{fig:MURIM_framework}
\end{figure}


\noindent \textbf{Resource Budget Verification.}
From the observed latency, the server infers each client’s effective resource usage $R_{i,t}$ and compares it with the declared budget $R_{i,t}^{\text{rep}}$. A belief band of width $\alpha_i$ marks consistency, while an extended band $\alpha_i+\epsilon_{R,i}$ permits small deviations. Values in the belief band set $b_{i,t}^{(R)}=1$, those in the extended band set $u_{i,t}^{(R)}=1$, and values outside both set $d_{i,t}^{(R)}=1$. With base rate $a=0.5$, the per-round probability of truthful reporting is
\begin{equation}
\label{eq:resource_prob}
P_{\text{resources},i,t}
= b_{i,t}^{(R)} + a\,u_{i,t}^{(R)}.
\end{equation}
Evidence is then accumulated across rounds to form reliability opinions.

\noindent \textbf{Privacy Budget Verification.}
We assess privacy behavior using an outlier-detection test on contributions $C_{i,t}$. Let $q_1$ and $q_3$ be the empirical quartiles and $\mathrm{IQR}=q_3-q_1$. The trusted interval is $[L,U]=[q_1-\kappa\,\mathrm{IQR},\,q_3+\kappa\,\mathrm{IQR}]$ with uncertainty margin $m=\gamma\,\mathrm{IQR}$. Scores in $[L,U]$ yield belief, those in $(L-m,L)\cup(U,U+m)$ yield uncertainty, and scores outside yield disbelief. With base rate $a=0.5$, we compute
\begin{equation}
P_{\text{privacy},i,t}=b_{i,t}^{(\varepsilon)}+a\,u_{i,t}^{(\varepsilon)}, \qquad
P_{\text{reliability},i,t}=P_{\text{resources},i,t}\,P_{\text{privacy},i,t}.
\end{equation}
Evidence accumulates across rounds to form long-term reliability opinions.

\noindent \textbf{Incentive Computation.}
Client incentives combine contribution, latency, and reliability scores:
\begin{equation}
I_{i,t}
= \mathbf{a}C_{i,t}
+ \mathbf{b}\,\frac{1}{L_{i,t}}
+ \mathbf{c}\,P_{\text{reliability},i,t}^{\zeta}
\!\left(1 + \exp\!\left[-\frac{P_{\text{reliability},i,t}-r_0}{s}\right]\right)^{-1}.
\label{eq:incentive-function}
\end{equation}
Here, $I_{i,t}$ is the incentive for client $i$ in round $t$, with $(\mathbf{a},\mathbf{b},\mathbf{c})$ weighting the three components; $r_0$ and $s$ tune the reliability transition, and $\zeta$ controls its influence. 

\noindent \textbf{Client Utility.}
Client utility is defined as
\begin{equation}
U_{i,t} = \frac{I_{i,t}\,\Omega - R_{i,t}}{\Omega},
\end{equation}
where $U_{i,t}$ is the net benefit for client $i$ in round $t$, $I_{i,t}$ is the server-issued incentive, and $R_{i,t}$ is the reported resource usage. The scaling factor $\Omega$ normalizes incentives and MAC-level resource costs so neither term dominates numerically.

Fig.~\ref{fig:MURIM_framework} provides a compact overview of the MURIM workflow, illustrating the interaction between client-side privacy-preserving training and server-side reputation, scoring, and reward mechanisms.

\noindent \textbf{Datasets.}
We evaluate on MNIST~\cite{lecun1998mnist}, FMNIST~\cite{xiao2017fmnist}, and ADULT~\cite{kohavi1996adult} in both IID and non-IID settings. MNIST and FMNIST are 10-class grayscale image benchmarks, while ADULT is a heterogeneous tabular dataset used to predict whether an individual’s income exceeds \$50K.

\noindent \textbf{Metrics.}
\textbf{(a) Accuracy} measures the final test performance of the aggregated global model.  
\textbf{(b) Client Utility} quantifies how well incentives align with each client’s resource expenditure.  
\textbf{(c) Attack Success Rate (ASR)} is used for privacy attacks, reported as \textbf{MIA ASR} and \textbf{PIA ASR}, where higher values indicate worse privacy.  
\textbf{(d) Accuracy Degradation} for poisoning attacks is computed as
$\Delta\text{Acc}=\text{Acc}_{\text{clean}}-\text{Acc}_{\text{attacked}}$,
reported for \textbf{MPA} and \textbf{NGA}, where larger positive values indicate stronger degradation.

\noindent \textbf{Baselines.}  For privacy, \textbf{DP-FedAvg}~\cite{geyer2017differentially} adds differential privacy, and \textbf{SecAgg}~\cite{bonawitz2017practical} performs secure aggregation.  For fairness and robustness, \textbf{FedAvg}~\cite{mcmahan2017communication} is standard averaging, \textbf{qFedAvg}~\cite{li2020fairresourceallocationfederated} upweights high-loss clients, \textbf{FedProx}~\cite{li2020federated} stabilizes client drift, and \textbf{Krum}~\cite{blanchard2017byzantine} selects majority-consistent updates for Byzantine resilience.  For training-time incentives, \textbf{IAFL}~\cite{wu2024incentive} and \textbf{FGCML}~\cite{xu2021gradient} reward clients via contribution signals.

\section{Experimental Results}
\noindent \textbf{Effect of Varying Reliability Thresholds.}
The reliability threshold is the minimum $P_{\text{reliability}}$ score required for a client to be treated as reliable, decreasing as a client deviates from expected behavior. Across datasets (Table~\ref{tab:varying_reliability_threshold}), IID settings show stable performance for all thresholds: accuracies remain nearly unchanged, no honest clients are removed, and all liars are filtered, indicating that reliability screening does not disrupt learning under homogeneous data. In non-IID settings, low thresholds yield few dropouts but permit more liars to survive, reducing accuracy. Increasing the threshold removes more unreliable clients and can improve performance (e.g., MNIST and FMNIST at mid-range values), but overly strict thresholds begin to drop honest clients, causing accuracy to decline. The ADULT non-IID dataset exhibits the greatest sensitivity to this threshold.

\begin{table}[t!]
\caption{\textbf{Performance vs.\ Reliability Threshold (100 Clients, Liar Fraction = 0.1).} 
Thr.\ denotes the reliability threshold; Train/Test/Val indicate accuracies; 
Innoc.\ Drop counts honest clients removed; Liar.\ Surv counts liars that survived; 
Dropouts represent the total number of client dropouts.}

\label{tab:varying_reliability_threshold}
\centering
\scriptsize
\begin{tabular}{|P{2cm}|c|c|c|c|c|c|c|}
\hline
\textbf{Dataset} & \textbf{Thr.} & \textbf{Test} & \textbf{Train} & \textbf{Val} & \textbf{Innoc.\ Drop} & \textbf{Liars Surv.} & \textbf{Dropouts} \\ \hline

\multirow{6}{*}{MNIST--IID}
 & 0.10 & 96.004  & 95.4972 & 96.116 & \textbf{0}   & \textbf{0}   & \textbf{10} \\ \cline{2-8}
 & 0.15 & 96.357  & 95.9862 & 96.577 & \textbf{0}   & \textbf{0}   & \textbf{10} \\ \cline{2-8}
 & 0.20 & 96.433  & 96.0444 & 96.726 & \textbf{0}   & \textbf{0}   & \textbf{10} \\ \cline{2-8}
 & 0.25 & \textbf{96.444}  & 96.0082 & 96.584 & \textbf{0}   & \textbf{0}   & \textbf{10} \\ \cline{2-8}
 & 0.30 & 96.254  & 95.9226 & 96.580 & \textbf{0}   & \textbf{0}   & \textbf{10} \\ \cline{2-8}
 & 0.35 & 96.153  & 95.7900 & 96.641 & \textbf{0}   & \textbf{0}   & \textbf{10} \\ \hline

\multirow{6}{*}{MNIST--nonIID}
 & 0.05 & 60.376  & 59.7438 & 61.504 & 0 & 1.6 & 8.4 \\ \cline{2-8}
 & 0.06 & 59.380  & 58.4768 & 59.915 & 0 & 1.4 & 8.6 \\ \cline{2-8}
 & 0.07 & 58.145  & 57.5878 & 59.337 & \textbf{0} & \textbf{1.0} & \textbf{9.0} \\ \cline{2-8}
 & 0.08 & \textbf{63.361}  & 61.9872 & 63.528 & 2.2 & 0.5 & 11.7 \\ \cline{2-8}
 & 0.09 & 59.215  & 57.8152 & 59.061 & 2.5 & 0.2 & 12.3 \\ \cline{2-8}
 & 0.10 & 57.731  & 56.9200 & 58.487 & 3.8 & 0.0 & 13.8 \\ \hline

\multirow{6}{*}{FMNIST--IID}
 & 0.10 & 74.468  & 75.1864 & 75.217 & \textbf{0}   & \textbf{0}   & \textbf{10} \\ \cline{2-8}
 & 0.15 & 74.461  & 75.3496 & 75.469 & \textbf{0}   & \textbf{0}   & \textbf{10} \\ \cline{2-8}
 & 0.20 & 74.281  & 75.0998 & 75.092 & \textbf{0}   & \textbf{0}   & \textbf{10} \\ \cline{2-8}
 & 0.25 & \textbf{74.966}  & 75.6824 & 75.778 & \textbf{0}   & \textbf{0}   & \textbf{10} \\ \cline{2-8}
 & 0.30 & 74.806  & 75.7374 & 75.813 & \textbf{0}   & \textbf{0}   & \textbf{10} \\ \cline{2-8}
 & 0.35 & 74.6926 & 75.5910 & 75.759 & \textbf{0}   & \textbf{0}   & \textbf{10} \\ \hline

\multirow{6}{*}{FMNIST--nonIID}
 & 0.05 & 46.347  & 46.5514 & 46.298 & 0 & 2.0 & 8.0 \\ \cline{2-8}
 & 0.06 & \textbf{51.045}  & 51.3360 & 51.028 & 0 & 1.6 & 8.4 \\ \cline{2-8}
 & 0.07 & 47.602  & 47.7808 & 47.793 & \textbf{0} & \textbf{0.6} & \textbf{9.4} \\ \cline{2-8}
 & 0.08 & 50.025  & 50.4056 & 50.398 & 1.2 & 0.1 & 11.1 \\ \cline{2-8}
 & 0.09 & 47.559  & 47.8462 & 47.585 & 2.7 & 0.2 & 12.5 \\ \cline{2-8}
 & 0.10 & 46.675  & 46.8888 & 47.014 & 3.7 & 0.0 & 13.7 \\ \hline

\multirow{6}{*}{ADULT--IID}
 & 0.10 & 74.2914 & 74.3007 & 74.1720 & \textbf{0} & \textbf{0} & \textbf{10} \\ \cline{2-8}
 & 0.15 & 68.3027 & 68.2745 & 68.0539 & \textbf{0} & \textbf{0} & \textbf{10} \\ \cline{2-8}
 & 0.20 & 70.3895 & 70.5703 & 70.3358 & \textbf{0} & \textbf{0} & \textbf{10} \\ \cline{2-8}
 & 0.25 & \textbf{74.8765} & 74.9378 & 74.8580 & \textbf{0} & \textbf{0} & \textbf{10} \\ \cline{2-8}
 & 0.30 & 67.7109 & 67.6248 & 67.5648 & \textbf{0} & \textbf{0} & \textbf{10} \\ \cline{2-8}
 & 0.35 & 69.4679 & 69.5643 & 69.5813 & \textbf{0} & \textbf{0} & \textbf{10} \\ \hline

\multirow{6}{*}{ADULT--nonIID}
 & 0.10 & 56.2805 & 56.1785 & 56.1140 & \textbf{0} & \textbf{1.4} & \textbf{8.6} \\ \cline{2-8}
 & 0.15 & 49.0154 & 49.0598 & 48.7523 & 1.6 & 0.3 & 11.3 \\ \cline{2-8}
 & 0.20 & 46.8971 & 47.0224 & 46.7959 & 5.1 & 0.0 & 15.1 \\ \cline{2-8}
 & 0.25 & \textbf{63.1477} & 63.2579 & 62.8560 & 7.6 & 0.0 & 17.6 \\ \cline{2-8}
 & 0.30 & 58.9176 & 58.7479 & 58.9202 & 10.0 & 0.0 & 20.0 \\ \cline{2-8}
 \hline

\end{tabular}
\end{table}

\begin{table}[t]
\caption{\textbf{Performance Across Client Counts for MNIST (Liar Fraction = 0.1).} 
Thr.\ denotes the reliability threshold; Train/Test/Val indicate accuracies; 
Innoc.\ Drop counts honest clients removed; Liar.\ Surv counts liars that survived; 
Dropouts represent the total number of client dropouts.}

\label{varying_num_clients}
\centering
\scriptsize
\begin{tabular}{|c|c|c|c|c|c|c|c|c|}
\hline
\textbf{Dataset} & \textbf{\#Clients} & \textbf{Thr.} & \textbf{Test} & \textbf{Train} & \textbf{Val} & \textbf{Innoc.\ Drop} & \textbf{Liars Surv.} & \textbf{Dropouts} \\ \hline

\multirow{6}{*}{MNIST--IID}
 & 10  & 0.25 & \textbf{98.803} & 99.0772 & 99.174 & \textbf{0} & \textbf{0} & \textbf{1} \\ \cline{2-9}
 & 50  & 0.25 & 97.193 & 96.9600 & 97.454 & \textbf{0} & \textbf{0} & \textbf{5} \\ \cline{2-9}
 & 100 & 0.25 & 96.554 & 96.2238 & 96.754 & \textbf{0} & \textbf{0} & \textbf{10} \\ \cline{2-9}
 & 200 & 0.25 & 92.219 & 91.4356 & 92.752 & \textbf{0} & \textbf{0} & \textbf{20} \\ \cline{2-9}
 & 300 & 0.25 & 91.544 & 90.7528 & 92.219 & \textbf{0} & \textbf{0} & \textbf{30} \\ \cline{2-9}
 & 500 & 0.25 & 91.078 & 90.2910 & 91.9418 & \textbf{0} & \textbf{0} & \textbf{50} \\ \hline

\multirow{6}{*}{MNIST--nonIID}
 & 10  & 0.07 & 59.715 & 58.5396 & 60.489 & 0 & 0.1 & 0.9 \\ \cline{2-9}
 & 50  & 0.07 & 55.100 & 52.8000 & 53.189 & 0 & 0.4 & 4.6 \\ \cline{2-9}
 & 100 & 0.07 & 57.654 & 56.6658 & 57.994 & 0 & 0.7 & 9.3 \\ \cline{2-9}
 & 200 & 0.07 & \textbf{67.014} & 65.4050 & 67.230 & 0 & 0.9 & 19.1 \\ \cline{2-9}
 & 300 & 0.07 & 64.574 & 63.3288 & 65.065 & 0 & 1.6 & 28.4 \\ \cline{2-9}
 & 500 & 0.07 & 65.194 & 63.8612 & 65.520 & 0 & 2.8 & 47.2 \\ \hline

\end{tabular}
\end{table}

\begin{table}[t!]
\caption{\textbf{FMNIST Performance vs.\ Liar Fraction (100 Clients).} 
Liar Frac.\ denotes the fraction of unreliable clients; Train/Test/Val indicate accuracies; 
Innoc.\ Drop counts honest clients removed; Liar.\ Surv counts liars that survived; 
Dropouts represent the total number of client dropouts.}
\centering
\scriptsize
\begin{tabular}{|l|c|c|c|c|c|c|c|}
\hline
\textbf{Dataset} & \textbf{Liar Frac.} & \textbf{Test} & \textbf{Train} & \textbf{Val} & \textbf{Innoc.\ Drop} & \textbf{Liars Surv.} & \textbf{Dropouts} \\ \hline
\label{varying_liar_frac}

\multirow{6}{*}{FMNIST--IID}
 & 0.50 & 74.495 & 75.3776 & 75.540 & \textbf{0} & \textbf{0} & \textbf{5}  \\ \cline{2-8}
 & 0.10 & 74.669 & 75.5290 & 75.559 & \textbf{0} & \textbf{0} & \textbf{10} \\ \cline{2-8}
 & 0.15 & \textbf{75.008} & 76.0166 & 76.158 & \textbf{0} & \textbf{0} & \textbf{15} \\ \cline{2-8}
 & 0.20 & 74.368 & 75.1226 & 75.119 & \textbf{0} & \textbf{0} & \textbf{20} \\ \cline{2-8}
 & 0.25 & 74.385 & 75.2310 & 75.251 & \textbf{0} & \textbf{0} & \textbf{25} \\ \cline{2-8}
 & 0.30 & 74.944 & 75.8818 & 75.917 & \textbf{0} & \textbf{0} & \textbf{30} \\ \hline

\multirow{6}{*}{FMNIST--nonIID}
 & 0.05 & 48.806 & 48.9790 & 48.908 & 0 & 0.3 & 4.7 \\ \cline{2-8}
 & 0.10 & 47.278 & 47.5442 & 47.504 & 0 & 0.7 & 9.3 \\ \cline{2-8}
 & 0.15 & 48.711 & 48.9026 & 48.790 & 0 & 1.4 & 13.6 \\ \cline{2-8}
 & 0.20 & 43.700 & 43.8670 & 43.902 & 0 & 1.2 & 18.8 \\ \cline{2-8}
 & 0.25 & 49.113 & 49.4796 & 49.218 & 0 & 2.6 & 22.4 \\ \cline{2-8}
 & 0.30 & \textbf{51.504} & 51.6474 & 51.630 & 0 & 2.0 & 28.0 \\ \hline

\end{tabular}
\end{table}

\noindent \textbf{Effect of Varying Number of Clients.}
Across client scales (Table~\ref{varying_num_clients}), the reliability mechanism consistently protects honest participants. In both IID and non-IID MNIST, no innocent clients are removed even as the number of clients grows from 10 to 500. Liar detection is perfect in IID settings, with zero survivors at all scales. In non-IID settings, surviving liars remain very few ($\leq 2.8$ at 500 clients) and increase only slightly as the system scales.
\\ \vspace{-4mm}\begin{wrapfigure}{r}{0.55\linewidth}
    \centering
    \vspace{-7mm}
    \includegraphics[width=\linewidth]{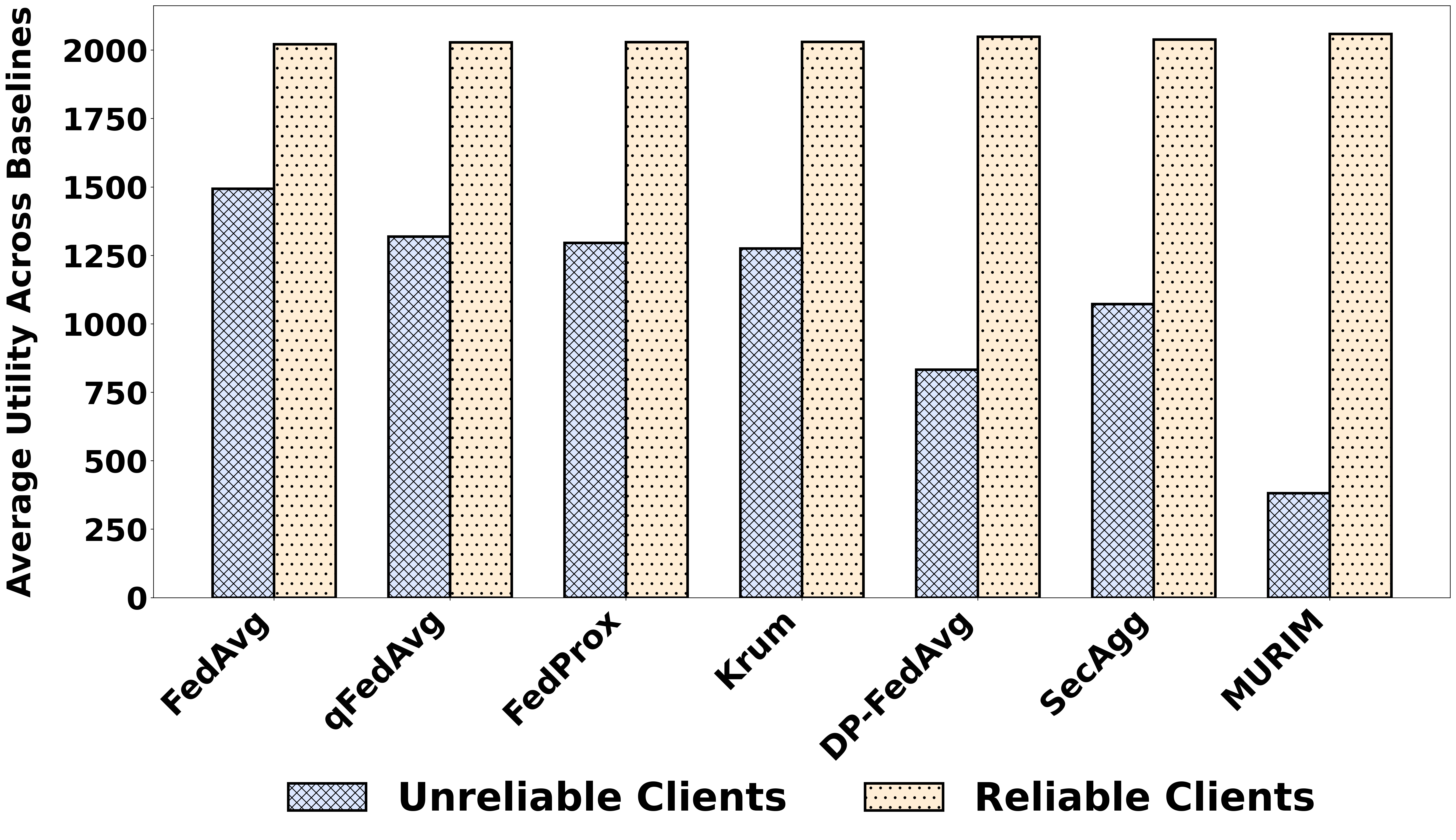}
    \vspace{-7mm}
    \caption{\bf Utilities of Reliable and Unreliable Clients on ADULT Non-IID}
    \label{fig:rel_vs_unrel_client_utilities}
    \vspace{-7mm}
\end{wrapfigure} 
\noindent \textbf{Effect of Varying Fraction of Unreliable Clients.}
Across liar fractions on FMNIST (Table~\ref{varying_liar_frac}), the mechanism consistently protects honest clients, with zero innocent clients dropped in both IID and non-IID settings, even as the liar fraction increases to 0.50 (IID) and 0.30 (non-IID).  In IID FMNIST, liar detection is perfect at all fractions, with no liars surviving and dropouts matching the injected liar count. In the more challenging non-IID case, surviving liars rise gradually with the liar fraction but remain small relative to the total, indicating that the mechanism remains highly effective under heterogeneity.

\noindent \textbf{Comparative Utility Analysis.}
Fig.~\ref{fig:rel_vs_unrel_client_utilities} compares the average utilities of reliable and unreliable clients across seven baselines. Reliable clients consistently receive higher utility under all methods, whereas unreliable clients earn substantially less. \murim{} exhibits the strongest separation, giving the highest utility to reliable clients and the lowest to unreliable ones, demonstrating effective discrimination between honest and dishonest behavior.

\begin{table}[h!]
\caption{\textbf{Privacy and Robustness Performance under Four Attacks.} 
MIA/PIA ASR ($\downarrow$) denotes attack success rates, while MPA/NGA $\Delta$Acc ($\downarrow$) represents accuracy degradation under model poisoning and noisy-gradient attacks.}
\label{tab:attack-results}
\centering
\scriptsize
\begin{tabular}{|c|c|c|c|c|}
\hline
& \multicolumn{2}{c|}{\bf Privacy Attack} & \multicolumn{2}{c|}{\bf Robustness Attack (pp)} \\ \hline
Method & MIA ASR ($\downarrow$) & PIA ASR ($\downarrow$) & MPA $\Delta$Acc ($\downarrow$) & NGA $\Delta$Acc ($\downarrow$) \\ \hline

FedAvg      & 47.13 & 44.69 & 1.29 & 0.28 \\ \hline
DP-FedAvg   & 47.97 & 45.23 & 5.92 & 1.81 \\ \hline
q-FedAvg    & 47.14 & 44.67 & 1.56 & 0.32 \\ \hline
FedProx     & 47.14 & 44.69 & 1.44 & 0.35 \\ \hline
SecAgg      & 47.15 & 44.69 & 1.60 & 1.14 \\ \hline
FGCML       & 47.73 & 47.04 & 2.55 & 0.30 \\ \hline
IAFL        & 48.48 & 51.98 & 2.07 & 1.72 \\ \hline
\textbf{MURIM} & \textbf{46.77} & \textbf{43.14} & \textbf{0.87} & \textbf{0.02} \\ \hline

\end{tabular}
\end{table}

\noindent \textbf{Adversarial Robustness Evaluation.}
\label{sec:attack-results}
Table~\ref{tab:attack-results} shows consistent trends across privacy and robustness metrics, where lower values indicate stronger defenses. \murim{} achieves the strongest results, with the lowest MIA (46.77) and PIA (43.14) ASRs and minimal poisoning degradation (MPA $\Delta$Acc 0.87 pp; NGA $\Delta$Acc 0.02 pp). In contrast, \textbf{DP-FedAvg} and \textbf{IAFL} incur larger robustness drops, and \textbf{IAFL} also shows higher PIA ASR. Classical baselines—\textbf{FedAvg}, \textbf{qFedAvg}, \textbf{FedProx}, and \textbf{SecAgg}, perform moderately. Overall, \murim{} more effectively couples privacy preservation with resilience to adversarial gradient perturbations while maintaining stability under noise.

\section{Conclusions \& Future Work}

\noindent \textbf{Conclusions.}
\murim{} delivers a unified, multidimensional incentive mechanism that jointly evaluates contribution quality, latency--resource consistency, privacy compliance, and long-term reliability to support fair, truthful, and privacy-preserving federated learning. By integrating Subjective Logic–based reliability assessment and the Subspace Leverage Equalizer (SLE), \murim{} addresses gaps left by existing mechanisms that treat fairness, privacy, or truthfulness in isolation. Empirical results across MNIST, FMNIST, and ADULT demonstrate that \murim{} reliably filters strategic clients, maintains stable convergence under heterogeneity, and achieves notable performance gains, including up to 18\% improvement in fairness metrics, 5--9\% reductions in privacy attack success rates, and as much as 85\% increased robustness against poisoning and noisy-gradient attacks. These findings show that \murim{} more effectively couples privacy preservation, incentive alignment, and resilience to adversarial behavior than classical, privacy-oriented, or incentive-based baselines.

\noindent \textbf{Limitations \& Future Work.}
While \murim{} demonstrates strong empirical performance, several limitations suggest natural extensions. The current evaluation focuses on standard vision and tabular benchmarks; applying \murim{} to larger and multimodal FL settings would further validate its scalability. Moreover, although \murim{} empirically ensures truthful reporting and liar detection, formal guarantees on incentive compatibility and robustness to coordinated misreporting or collusion remain to be developed. Additionally, the framework uses fixed privacy-budget intervals and resource expectations; future work will investigate adaptive, personalized contracts that dynamically adjust these parameters in response to evolving client behavior. Finally, deploying \murim{} in real-world cross-device and cross-silo systems will reveal practical constraints and guide refinements that enhance its reliability under operational conditions.

\bibliographystyle{splncs04}
\bibliography{main.bib}
\end{document}